\documentclass{article}
\usepackage{academic-blue}
\usepackage[T1]{fontenc}
\usepackage{amsmath,amssymb,booktabs,array,tabularx,multirow,longtable}
\usepackage{colortbl}
\usepackage{float}
\usepackage{placeins}

\newcolumntype{Y}{>{\raggedright\arraybackslash}X}
\renewenvironment{abstract}{%
  \par\begingroup\small
  \noindent\textcolor{DeepBlue}{\textbf{Abstract}}\quad\ignorespaces
}{\par\endgroup}
\title{\normalfont\bfseries\color{black}\textcolor[HTML]{347DBE}{\textit{VisualArtifactSpec 1.0:}}\\Evaluating Visual Artifacts under Task Specifications}
\author{%
  \textbf{Shuyi Ouyang\textsuperscript{1,2*}\quad
  Xiaojun Tang\textsuperscript{2*}\quad
  Zijian Zou\textsuperscript{2}\quad
  Xintong Zhang\textsuperscript{2}}\\
  \textbf{Mingli Song\textsuperscript{1}\quad
  Guannan Zhang\textsuperscript{2\textdagger}\quad
  Hongwei Xue\textsuperscript{2\textdagger}}\\[4pt]
  \normalfont\textsuperscript{1}Zhejiang University\\
  \normalfont\textsuperscript{2}Accio Team, Alibaba Group\\[3pt]
  {\normalfont\small\textsuperscript{*}Equal contribution\quad
  \textsuperscript{\textdagger}Corresponding author}
}
\headerlogos{\includegraphics[height=0.5cm]{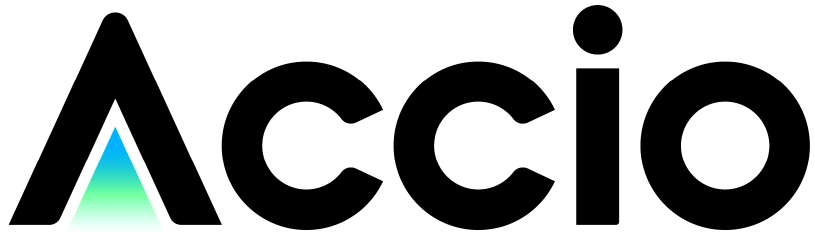}}
\hypersetup{pdftitle={VisualArtifactSpec 1.0: Evaluating Visual Artifacts under Task Specifications},pdfauthor={Shuyi Ouyang, Xiaojun Tang, Zijian Zou, Xintong Zhang, Mingli Song, Guannan Zhang, Hongwei Xue}}
\begin{document}
\raggedbottom
\maketitle
\vspace{-14pt}
\begin{abstract}
Visual artifact tasks test agents' ability to translate task specifications and source materials into outputs that meet practical needs. Evaluation across artifact families helps distinguish broadly applicable authoring capabilities from strengths confined to a particular medium. We introduce VisualArtifactSpec 1.0 (VAS), a benchmark of 211 creation and modification tasks spanning presentations, static graphics, animations and videos, and web interfaces. Each task specifies authoring objectives, source materials, constraints, and deliverables. Existing benchmarks provide limited support for systematically evaluating how authoring choices suit different audiences, purposes, and media across artifact families. VAS evaluates two aspects separately: Task Fulfillment verifies explicit requirements and applicable functionality, while Artifact Performance assesses Content, Organization, Visuals, and Delivery. Task-specific rating anchors instantiate shared criteria for the intended audience, purpose, and medium, with judgments grounded in source materials and artifact inspection records. We evaluate nine model--harness configurations on all 211 tasks. Overall scores range from 42.12 to 70.52, with Artifact Performance accounting for 73.4\% of the gap between the highest- and lowest-scoring configurations.
\end{abstract}
\nobreak\vspace{3pt}
\begingroup
\normalfont\fontsize{8.5}{10.5}\selectfont
\renewcommand{\arraystretch}{1.15}
\noindent\begin{tabular}{@{}l@{}}
  \makebox[1.5em][l]{\raisebox{-1.2pt}{\includegraphics[height=1em]{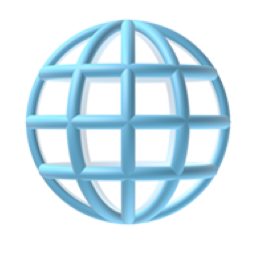}}}\textbf{Project Page:} \href{https://accio-lab.github.io/VisualArtifactSpec/}{https://accio-lab.github.io/VisualArtifactSpec}\\
  \makebox[1.5em][l]{\raisebox{-1.2pt}{\includegraphics[height=1em]{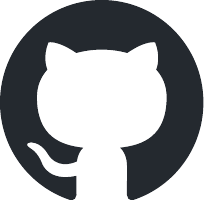}}}\textbf{GitHub Repo:} \href{https://github.com/Accio-Lab/VisualArtifactSpec}{https://github.com/Accio-Lab/VisualArtifactSpec}\\
  \makebox[1.5em][l]{\raisebox{-1.2pt}{\includegraphics[height=1em]{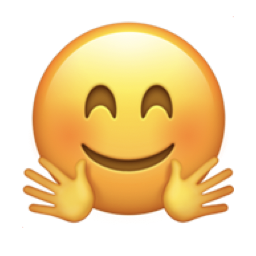}}}\textbf{HuggingFace:} \href{https://huggingface.co/datasets/Accio-Lab/VisualArtifactSpec}{https://huggingface.co/datasets/Accio-Lab/VisualArtifactSpec}
\end{tabular}\par
\endgroup
\vfill
\noindent\begin{minipage}{\linewidth}
\centering
\includegraphics[width=\linewidth]{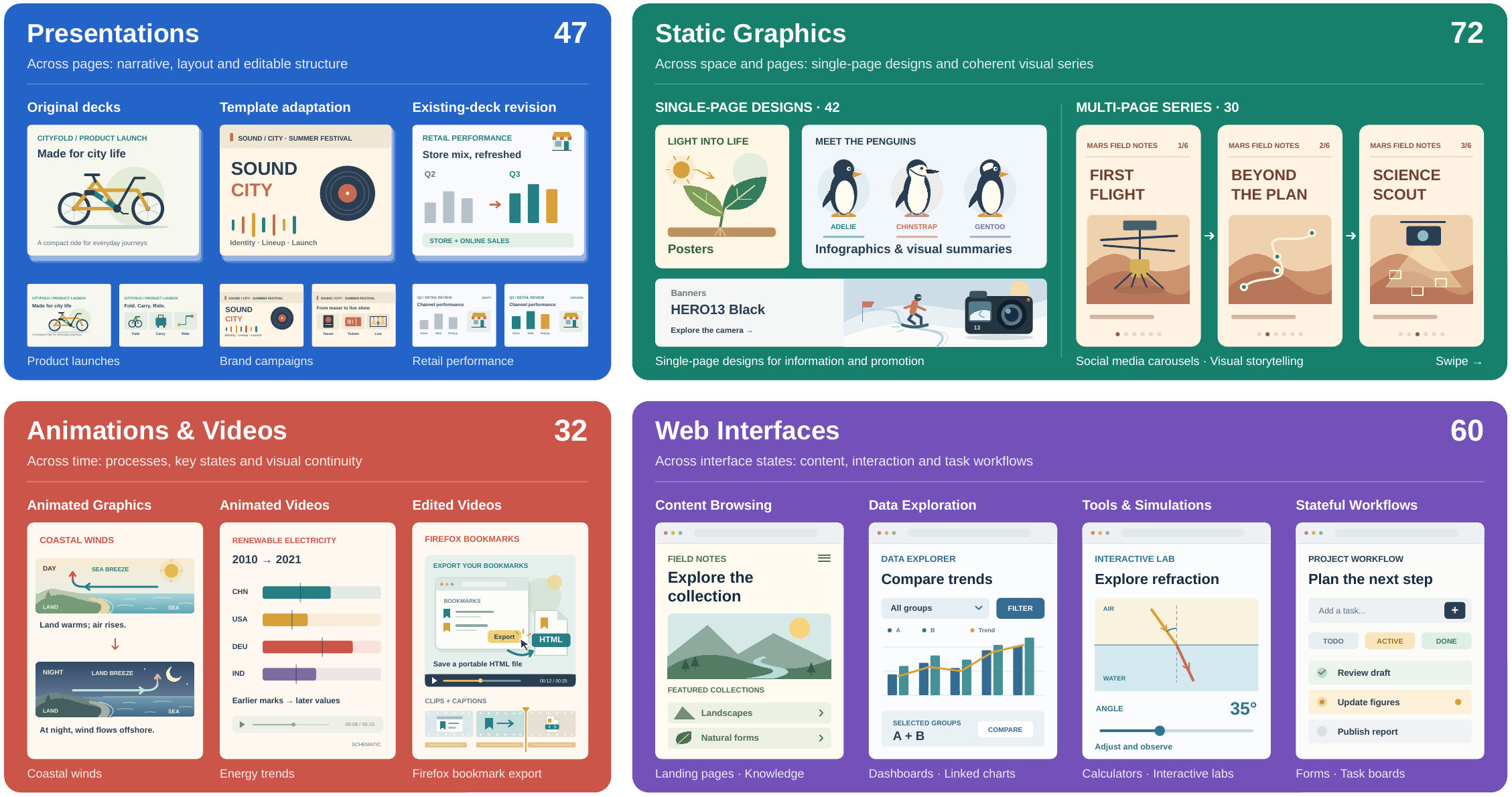}
\captionsetup{skip=5pt,hypcap=false}
\captionof{figure}{Representative visual artifacts from the four families in VAS.}
\label{fig:overview}
\end{minipage}\par
\clearpage
\suppressfloats[t]
\section{Introduction}
Visual artifacts make an agent's interpretation of a task concrete and inspectable. Presentations must develop an account across slides, animations must convey changes over time, and web interfaces must support browsing and interaction. These tasks share demands for source integration, content organization, and visual communication, but place different constraints on their execution. Evaluating multiple artifact families therefore helps determine whether an agent's strengths extend across deliverable forms or remain concentrated in a particular medium.

Table~\ref{tab:benchmark-comparison-candidate} compares the artifact coverage and evaluation criteria of existing benchmarks. PresentBench \citep{chen2026presentbench}, Code2Video \citep{chen2025code2video}, and Vision2Web \citep{he2026vision2web} establish detailed protocols for presentations, educational videos, and websites, respectively. ArtifactsBench extends evaluation to multiple executable artifact types, including checks of instruction compliance, functionality, visual presentation, and content accuracy \citep{zhang2025artifactsbench}. These advances leave a further evaluation need: systematically assessing how authoring choices serve different audiences, purposes, and media under a common framework across artifact families. Broad coverage, requirement satisfaction, and content accuracy alone do not establish whether the selected content, structure, and presentation are appropriate for the intended context.

\begin{table}[!htbp]
\centering
\setlength{\tabcolsep}{3pt}
\renewcommand{\arraystretch}{1.15}
\definecolor{CandidateCoverage}{HTML}{F3EFE7}
\definecolor{CandidateTF}{HTML}{EAF1F7}
\definecolor{CandidateAP}{HTML}{EBF3EE}
\definecolor{CandidateDashGray}{HTML}{858585}
\newcommand{\candidateDash}{\textcolor{CandidateDashGray}{--}}
\newcommand{\candidateCheck}{\textcolor{black}{\ensuremath{\checkmark}}}
\newcommand{\candidateSubHead}[1]{{\fontsize{7.5}{9}\selectfont\mdseries #1}}
\newcommand{\candidateCoverageText}[1]{{\fontsize{7}{9}\selectfont #1}}
\newcommand{\candidateGroupHead}[1]{\shortstack[c]{\strut\textbf{#1}}}
\caption{Artifact coverage and evaluation criteria of representative benchmarks. For evaluation criteria, \candidateCheck{}: routinely evaluated; \candidateDash{}: routine evaluation not established. Definitions and evidence are provided in Appendix~\ref{app:benchmark-comparison}.}
\label{tab:benchmark-comparison-candidate}
\fontsize{8}{10}\selectfont
\begin{tabularx}{\linewidth}{@{}Y>{\columncolor{CandidateCoverage}[\tabcolsep][1.5pt]}c>{\columncolor{CandidateTF}[1.5pt][\tabcolsep]}c>{\columncolor{CandidateTF}[\tabcolsep][1.5pt]}c>{\columncolor{CandidateAP}[1.5pt][\tabcolsep]}c*{2}{>{\columncolor{CandidateAP}}c}>{\columncolor{CandidateAP}[\tabcolsep][0pt]}c@{}}
\toprule
\rowcolor{white}
\multicolumn{1}{@{}l}{\candidateGroupHead{Benchmark}} & \candidateGroupHead{Coverage} & \multicolumn{2}{c}{\candidateGroupHead{Task Fulfillment}} & \multicolumn{4}{c@{}}{\candidateGroupHead{Artifact Performance}}\\
\cmidrule(lr){3-4}\cmidrule(lr){5-8}
\rowcolor{white}
 & & \candidateSubHead{Instruction} & \candidateSubHead{Functional} & \candidateSubHead{Visual} & \candidateSubHead{Content} & \candidateSubHead{Audience} & \candidateSubHead{Medium}\\[-2pt]
\rowcolor{white}
 & & \candidateSubHead{compliance} & \candidateSubHead{verification} & \candidateSubHead{presentation} & \candidateSubHead{accuracy} & \candidateSubHead{alignment} & \candidateSubHead{adaptation}\\
\midrule
StructBench \citep{zhuo2026structbench} & \candidateCoverageText{G} & \candidateCheck{} & \candidateDash{} & \candidateCheck{} & \candidateCheck{} & \candidateDash{} & \candidateDash{}\\
DrawAI-Bench \citep{cao2026drawai} & \candidateCoverageText{G} & \candidateCheck{} & \candidateDash{} & \candidateCheck{} & \candidateCheck{} & \candidateDash{} & \candidateDash{}\\
BizGenEval \citep{li2026bizgeneval} & \candidateCoverageText{G} & \candidateCheck{} & \candidateDash{} & \candidateCheck{} & \candidateCheck{} & \candidateDash{} & \candidateDash{}\\
VAO-Bench \citep{guo2026vao} & \candidateCoverageText{G} & \candidateCheck{} & \candidateDash{} & \candidateCheck{} & \candidateCheck{} & \candidateDash{} & \candidateDash{}\\
PresentBench \citep{chen2026presentbench} & \candidateCoverageText{P} & \candidateCheck{} & \candidateDash{} & \candidateCheck{} & \candidateCheck{} & \candidateCheck{} & \candidateDash{}\\
Code2Video \citep{chen2025code2video} & \candidateCoverageText{M} & \candidateDash{} & \candidateDash{} & \candidateCheck{} & \candidateCheck{} & \candidateDash{} & \candidateDash{}\\
Vision2Web \citep{he2026vision2web} & \candidateCoverageText{W} & \candidateCheck{} & \candidateCheck{} & \candidateCheck{} & \candidateDash{} & \candidateDash{} & \candidateDash{}\\
EvoPresent \citep{liu2025evopresent} & \candidateCoverageText{P/M} & \candidateDash{} & \candidateDash{} & \candidateCheck{} & \candidateCheck{} & \candidateDash{} & \candidateDash{}\\
GDB \citep{deganutti2026graphicdesignbench} & \candidateCoverageText{G/M} & \candidateCheck{} & \candidateDash{} & \candidateCheck{} & \candidateDash{} & \candidateDash{} & \candidateDash{}\\
ArtifactsBench \citep{zhang2025artifactsbench} & \candidateCoverageText{G/M/W} & \candidateCheck{} & \candidateCheck{} & \candidateCheck{} & \candidateCheck{} & \candidateDash{} & \candidateDash{}\\
\textbf{VAS 1.0} & \candidateCoverageText{P/G/M/W} & \candidateCheck{} & \candidateCheck{} & \candidateCheck{} & \candidateCheck{} & \candidateCheck{} & \candidateCheck{}\\
\bottomrule
\end{tabularx}
\par\vspace{4pt}
\begin{minipage}{\linewidth}\fontsize{8}{10}\selectfont
\textit{Coverage.} P: presentation decks; G: static graphics; M: animations/videos; W: web interfaces. Slide images, single-slide reconstruction, and static webpage renderings do not count toward P or W.
\par\vspace{1pt}
\textit{Audience alignment}: fit to audience knowledge and needs; \textit{Medium adaptation}: fit to medium or platform usage preferences.
\end{minipage}
\end{table}

We introduce \textbf{VisualArtifactSpec 1.0 (VAS)}, a benchmark of 211 creation and modification tasks spanning presentations, static graphics, animations and videos, and web interfaces. Figure~\ref{fig:overview} shows representative artifacts from these four families. Task construction pairs source materials with authoring objectives, constraints, and required deliverables, while specifying the audience, purpose, and medium that inform both authoring and evaluation.

\begin{figure}[!htbp]
\centering
\includegraphics[width=\linewidth]{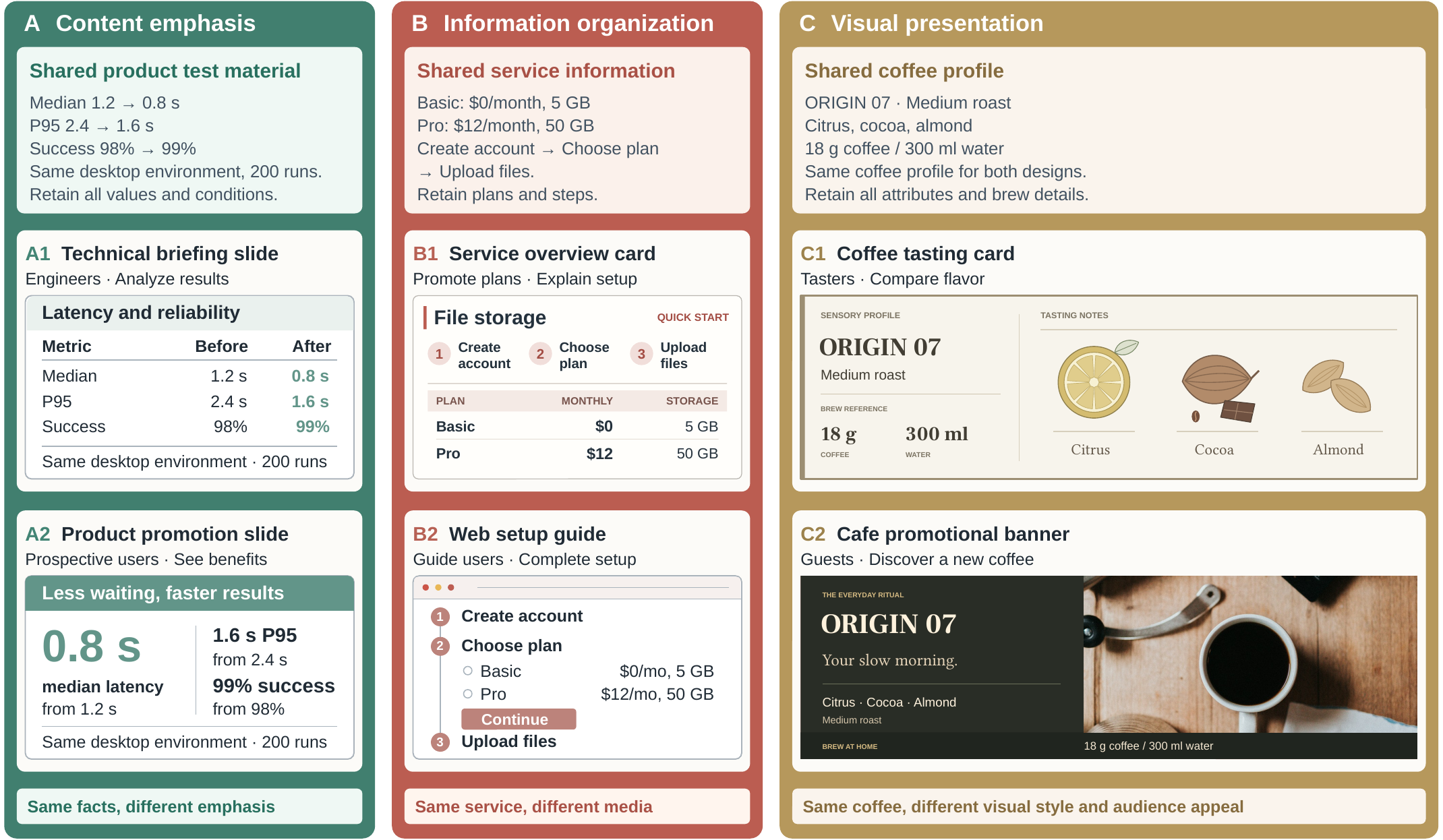}
\caption{Content emphasis (A), information organization (B), and visual presentation (C) can vary while required information is retained. The paired examples illustrate authoring choices shaped by audience, purpose, and medium.}
\label{fig:introduction-context}
\end{figure}

VAS evaluates two aspects separately: \emph{Task Fulfillment} verifies explicit requirements and applicable functionality, while \emph{Artifact Performance} assesses Content, Organization, Visuals, and Delivery. Shared criteria are instantiated through task-specific rating anchors informed by the audience, purpose, and medium. Figure~\ref{fig:introduction-context} illustrates why these anchors matter: with the required information retained, content emphasis can support technical analysis or product promotion (A), information organization can provide a compact service overview or stepwise web guidance (B), and visual presentation can support flavor comparison or lifestyle appeal (C). Judgments link the criteria to source materials and native, rendered, temporal, or interaction evidence, making unmet requirements distinguishable from weaknesses in authoring choices.

Experiments with nine model--harness configurations show that high Task Fulfillment scores can coexist with deficiencies in Artifact Performance. The cross-family comparisons in Section~\ref{sec:family-context-results} reveal strengths and weaknesses obscured by overall scores. The case studies in Section~\ref{sec:case-studies} connect score differences to evidence organization, mobile readability, and temporal consistency in the actual artifacts.

Our contributions are summarized as follows:
\vspace{-7pt}
\begin{itemize}
\item We construct VAS, a benchmark of 211 source-grounded creation and modification tasks across four artifact families, with explicit specifications of objectives, source materials, constraints, and deliverables.
\vspace{-6pt}
\item We develop an evidence-based evaluation protocol that separates Task Fulfillment from Artifact Performance, with task-specific rating anchors that instantiate shared Content, Organization, Visuals, and Delivery criteria for the intended audience, purpose, and medium.
\vspace{-6pt}
\item We compare nine model--harness configurations across artifact families and connect their fulfillment and performance differences to observable authoring choices through artifact-based case studies.
\end{itemize}
\FloatBarrier

\section{Related Work}
\subsection{Visual Artifact Authoring and Benchmarks}
GDPval evaluates professional work products, while AnalystBench studies reports grounded in multimodal sources \citep{patwardhan2025gdpval,pham2026analystbench}.

For presentations, AutoPresent introduces SlidesBench; PPTAgent assesses content, design, and coherence; and DeepPresenter refines rendered outputs \citep{ge2025autopresent,zheng2025pptagent,zheng2026deeppresenter}. PresentBench provides material-grounded checklists, while SlidesGen-Bench evaluates content, aesthetics, and editability \citep{chen2026presentbench,yang2026slidesgenbench}.

For graphics, ChartMimic evaluates chart-to-code generation, and Paper2Poster probes information conveyed by scientific posters \citep{yang2024chartmimic,pang2025paper2poster}. IGenBench and StructBench assess factual reliability, while BizGenEval checks commercial-image constraints \citep{tang2026igenbench,zhuo2026structbench,li2026bizgeneval}. Image-level checks do not establish native-file editability or executable behavior. DrawAI explicitly evaluates reconstruction fidelity and editability \citep{cao2026drawai}; VAO-Bench assesses edits, dependent updates, and preservation through an image-only protocol \citep{guo2026vao}. Graphic-Design-Bench covers layout, typography, infographics, design semantics, and animation \citep{deganutti2026graphicdesignbench}.

For animations and videos, Code2Video evaluates educational videos, while EvoPresent studies academic decks and presentation videos \citep{chen2025code2video,liu2025evopresent}. Animation2Code evaluates appearance and temporal similarity; design-video metrics cover layout, motion, temporal quality, and content fidelity \citep{ji2026animation2code,deganutti2026designvideo}.

For websites, Design2Code evaluates screenshot reconstruction, and DesignBench covers generation, editing, and repair \citep{si2024design2code,xiao2025designbench}. Vision2Web tests static, interactive, and full-stack development; LongWebBench evaluates long-page structure and interactions \citep{he2026vision2web,zhao2026longwebbench}.

Across families, ArtifactsBench evaluates executable visual code using source code, temporal screenshots, and task-specific checklists \citep{zhang2025artifactsbench}. VAS covers source-grounded creation and modification of both native and executable deliverables across four artifact families.

\begin{table}[t]
\centering\small
\caption{Task coverage in the VAS 1.0 release. Shares use 211 tasks as the denominator. Secondary categories are meaningful within each family and need not represent the same classification axis across families.}
\label{tab:coverage}
\begin{tabularx}{\linewidth}{@{}lrrY@{}}
\toprule Family & Tasks & Share & Task forms \\
\midrule
Presentations & 47 & 22.3\% & Original decks (31); template adaptation (14); existing-deck revision (2) \\
Static graphics & 72 & 34.1\% & Posters (8); infographics and visual summaries (24); banners (10); multi-page series (30) \\
Animations and videos & 32 & 15.2\% & Animated graphics (8); animated videos (22); edited videos (2) \\
Web interfaces & 60 & 28.4\% & Content browsing (22); data exploration (16); tools and simulations (13); stateful workflows (9) \\
\midrule Total & 211 & 100.0\% & Four artifact families \\
\bottomrule
\end{tabularx}
\end{table}

\begin{figure}[!t]
\centering
\includegraphics[width=0.86\linewidth]{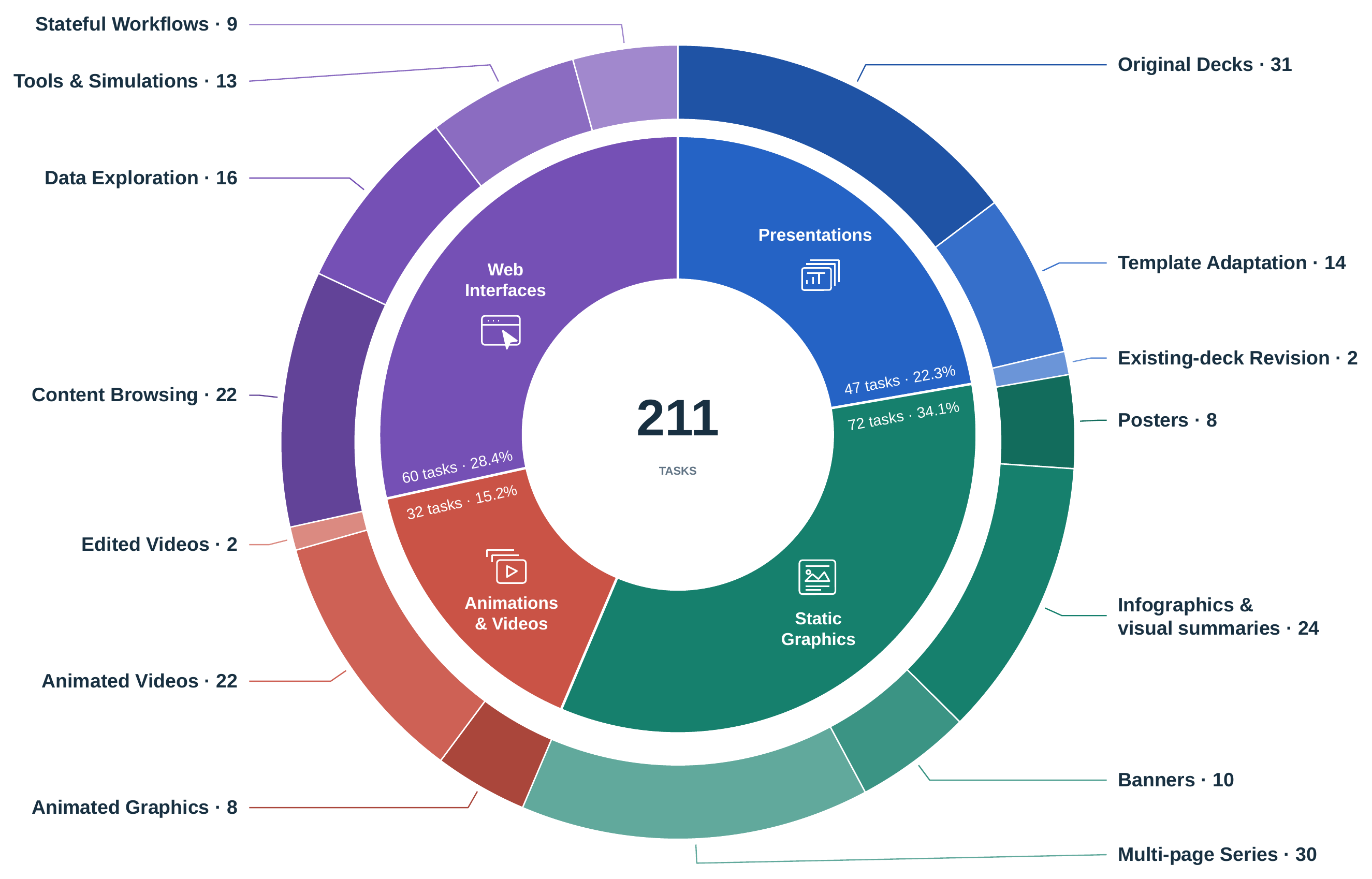}
\caption{Artifact families and 14 task types across 211 tasks, using the taxonomy illustrated in the overview. Task types follow task instructions and principal deliverables. Percentages use 211 as the denominator.}
\label{fig:artifact-types}
\end{figure}

\subsection{Rubric-Based Evaluation}
Criterion design ranges from instance-specific quality criteria in BiGGen Bench to hierarchical requirement decomposition in PaperBench \citep{kim2024biggen,starace2025paperbench}. In visual authoring, PresentBench aggregates binary judgments on instance-specific checklists, while ArtifactsBench scores task-specific criteria covering functionality and visual quality \citep{chen2026presentbench,zhang2025artifactsbench}. VAS separates binary Task Fulfillment checks from graded Artifact Performance ratings. Across artifact families, task-specific rating anchors connect shared Content, Organization, Visuals, and Delivery criteria to the intended audience, purpose, and medium, with judgments grounded in artifact evidence.

Judge validation addresses a different question. MLLM-as-a-Judge and Judge Anything compare multimodal judgments with human assessments; RubricEval examines criterion-level accuracy in instruction following \citep{chen2024mllmjudge,pu2025judgeanything,pan2026rubriceval}.

\begin{figure}[t]
\centering
\includegraphics[width=\linewidth]{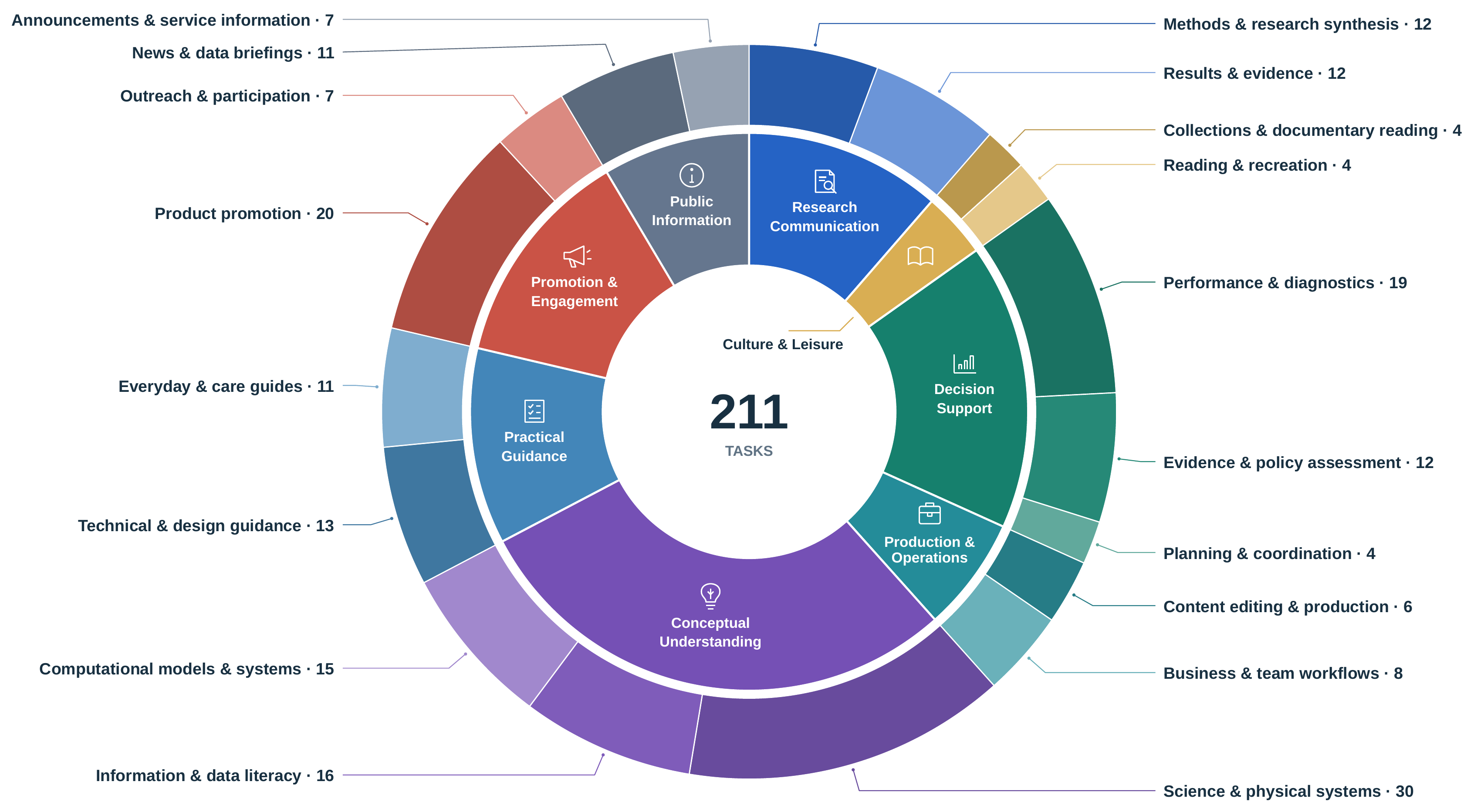}
\caption{Application purposes across 211 tasks. The inner ring shows primary purposes and the outer ring shows specific uses. Each task is counted once; percentages use 211 as the denominator.}
\label{fig:purposes}
\end{figure}

\section{VisualArtifactSpec 1.0}
\subsection{Task Definition and Coverage}
Each VAS task specifies an authoring objective, its intended audience, purpose, and medium, source materials, constraints, execution conditions, and required deliverables. Agents receive the instructions and inputs; evaluator-only materials provide reference facts, expected states, requirement checks, and rating anchors. Tasks cover creation, adaptation, revision, repair, and conversion, leaving content organization and presentation open within the stated requirements and modification boundaries.

\paragraph{Artifact families.}
Table~\ref{tab:coverage} summarizes the 211 tasks, and Figure~\ref{fig:artifact-types} shows their hierarchical distribution. Presentations develop an account across slides; static graphics organize information on a single page or across a series; animations and videos develop content over time; web interfaces support browsing, exploration, tools, and workflows. Classification follows the principal deliverable objective rather than the file extension: HTML can encode slides, an animation, or a website. Appendix~\ref{app:classification} gives classification rules, formats, and task identifiers.

\paragraph{Purposes and audiences.}
Figure~\ref{fig:purposes} groups the tasks into eight application purposes, each describing what the artifact should accomplish for its users. Every task receives one primary purpose and a more specific use; Appendix~\ref{app:scope} provides the mapping scope. Audience knowledge and needs shape the expected explanations, while medium and purpose shape information density, pacing, and presentation. For example, research communication emphasizes evidence and its limits, whereas practical guidance emphasizes actions and consequences. These distinctions inform task-specific evaluation criteria.

\subsection{Task Construction}
\begin{figure}[t]
\centering
\includegraphics[width=\linewidth]{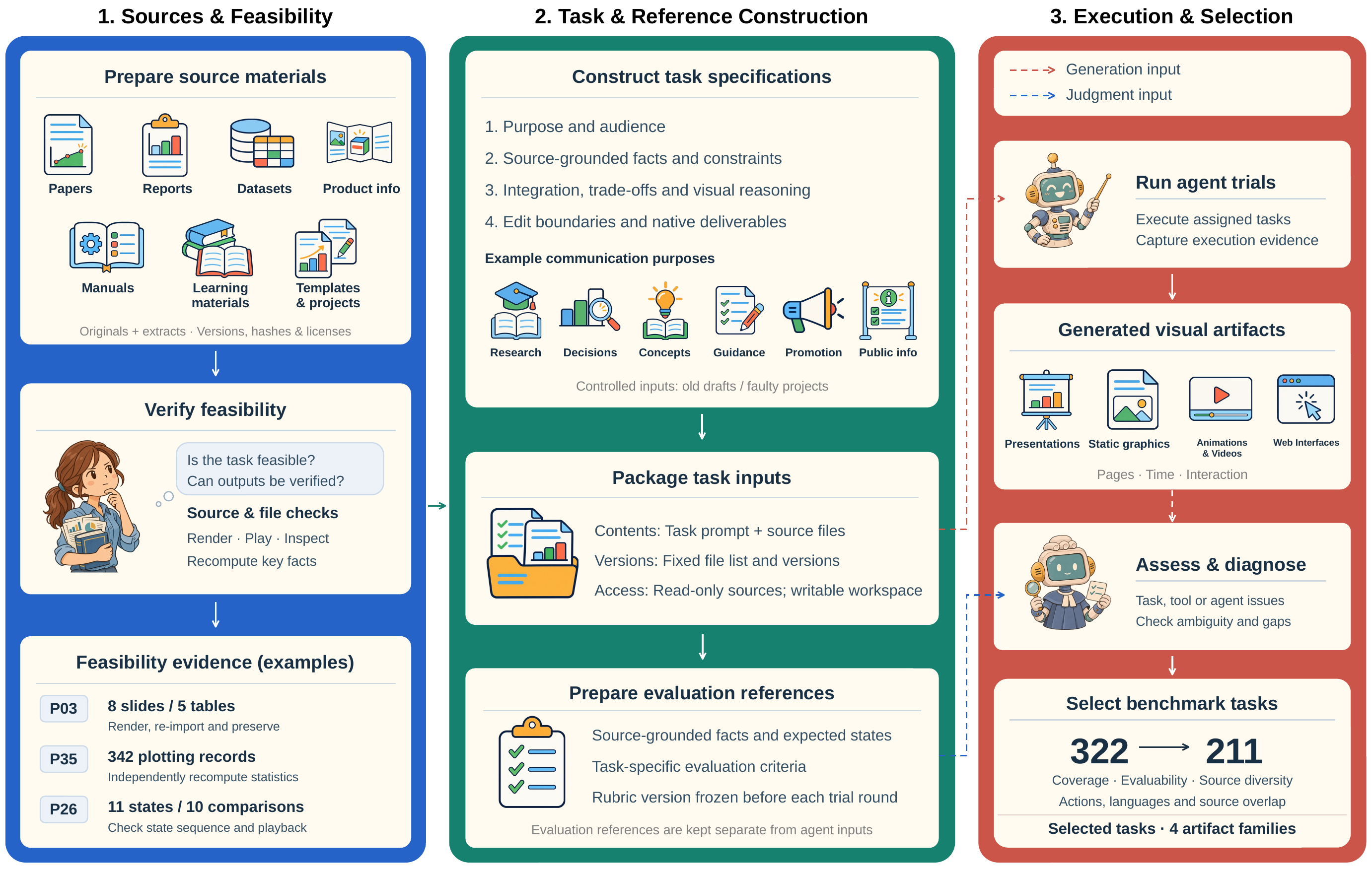}
\caption{Benchmark construction in three stages: sources and feasibility; task and reference construction; and execution and selection. Feasibility checks and communication purposes are illustrative examples. The selection step retains 211 release tasks from 322 candidates, considering coverage, evaluability, and source diversity.}
\label{fig:construction}
\end{figure}

Figure~\ref{fig:construction} summarizes three stages that produced 211 tasks from 322 candidates.

\paragraph{Stage 1: Sources and feasibility.}
We assemble public papers, reports, datasets, manuals, and media around concrete application objectives, retaining original files and source versions. Feasibility checks use rendering, native-file inspection, reproducible calculations, and temporal or interaction checks to establish usable inputs and observable outcomes. Controlled inputs supplement public sources for revision and repair: P03 supplies an earlier report grounded in official disclosures, and P26 supplies an animation with deliberate defects.

\paragraph{Stage 2: Task and reference construction.}
We translate source facts and application needs into instructions, input contracts, reference facts, and task-specific rubrics. Input contracts distinguish supplied files from writable outputs; references record factual checks, expected states, and acceptable modification boundaries. Package checks verify required files, manifests, paths, and rubric structure. Source checks resolve discrepancies against originals and independently recompute required quantities; for example, P03 checks cover financial values, native tables, and slides that must remain unchanged. Rubric revisions are versioned with their task packages.

\paragraph{Stage 3: Execution and selection.}
Agent trials expose ambiguous instructions, missing resources, observation failures, and artifact defects. Their results inform task revisions and checks of difficulty and score variation. Selection balances artifact and application coverage, evaluability, and source diversity. Same-source candidates are compared by their required operations and deliverables: creation and revision from the same materials can test different capabilities, whereas superficial variants add little coverage. Subsequent additions address gaps in artifact forms, authoring operations, and language coverage.

\subsection{Evaluation Framework}
\label{sec:evaluation-protocol}
\begin{figure}[!t]
\centering
\includegraphics[width=\linewidth]{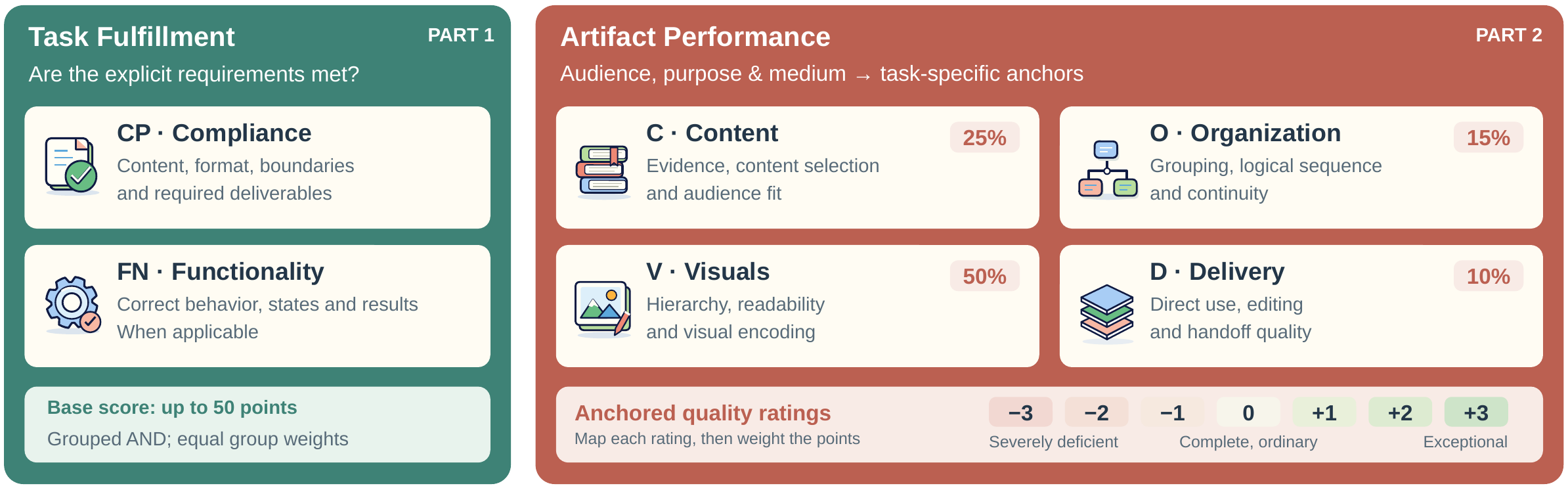}
\caption{The two-part evaluation rubric. Task Fulfillment allocates 50 points equally across acceptance groups, with AND within each group. Artifact Performance maps each rating to points before weighted aggregation, yielding an adjustment from $-50$ to $+50$. The total is the base plus this adjustment, without clipping.}
\label{fig:rubric-evaluation}
\end{figure}

VAS separates \textbf{Task Fulfillment}---Compliance (CP) and applicable Functionality (FN)---from \textbf{Artifact Performance}---Content (C), Organization (O), Visuals (V), and Delivery (D). The former verifies specified obligations; the latter assesses authoring quality through task-specific criteria within shared dimensions. Figure~\ref{fig:rubric-evaluation} summarizes the framework.

\paragraph{Requirement checks.}
CP verifies explicit requirements such as required facts, formats, modification boundaries, and native editability. FN verifies required operations and observable outcomes, including state transitions and consistency after interaction. In the current collection, FN applies to the 60 web-interface tasks; the other 151 tasks use CP for requirement checks. CP and FN items form acceptance groups, each assigned to one dimension and fixed before agent comparisons.

\paragraph{Performance criteria and contextual anchors.}
The 18 shared criteria comprise Content (3), Organization (3), Visuals (9), and Delivery (3), covering conceptual expression and evidence, structure and continuity, visual presentation, and usability and handoff. Table~\ref{tab:quality-criteria} defines their scopes. Task-specific anchors translate each scope into observable qualities for the intended audience, purpose, and medium. Table~\ref{tab:contextual-anchors} illustrates this for V8. The distinction from requirement checks is explicit: in P35, CP verifies required samples and calculations, while C1 assesses how clearly the graphic explains means, valid sample counts, and observational limits to first-year students.

\begin{table}[t]
\centering\small
\caption{Task-specific observation targets for V8 (scanning and temporal rhythm).}
\label{tab:contextual-anchors}
\setlength{\tabcolsep}{4pt}
\begin{tabularx}{\linewidth}{@{}lp{0.28\linewidth}Y@{}}
\toprule Task & Intended context & Observation targets \\
\midrule
P80 & Research-group slides & Within-page scanning order and clear stopping points at key conclusions across eight projected slides. \\
P99 & Public-facing mobile carousel & Entry points, density, and reading pace across five cards, including dense tables and image captions. \\
\bottomrule
\end{tabularx}
\end{table}

Each criterion receives an integer rating from $-3$ to $+3$. Negative ratings distinguish severe failure ($-3$), sustained impairment of major content or processes ($-2$), and localized substantive defects ($-1$). Zero denotes complete basic satisfaction without substantive defects, but ordinary quality. A rating of $+1$ denotes excellent execution with observable room for improvement; $+2$ requires full refinement across the applicable scope, with no substantive weakness or feasible task-relevant improvement of material benefit. A rating of $+3$ additionally requires a non-obvious design that resolves a real task challenge and yields substantial evidenced benefits beyond an equally correct, complete, and polished conventional solution. Appendix~\ref{app:rating-evidence} details rating evidence and item boundaries.

\paragraph{Evidence and judgment.}
Evaluation freezes the submitted artifact and its file manifest, then combines source materials with rendered views, native-file inspection, temporal samples, and interaction records as required. Websites are inspected at $1440\times900$ and $390\times844$ viewports and through required actions. A multimodal judge records item-level judgments with supporting pages, regions, timestamps, or interaction states, together with artifact, rubric, and judge versions. Unresolved evaluation failures remain unscored; Appendix~\ref{app:acceptance-groups} specifies missing-artifact and website-failure handling. Section~\ref{sec:experimental-setup} gives the judge configuration.

\subsection{Scoring and Reporting}
\label{sec:scoring-reporting}
Task Fulfillment allocates 50 points equally across acceptance groups, with \textbf{AND within each group}. Let $a_i\in\{0,1\}$ denote a resolved No or Yes judgment and $G_j$ the items in group $j$. For $m$ groups spanning CP and applicable FN,
\begin{equation}
g_j=\prod_{i\in G_j}a_i,
\qquad B=50\,\frac{\sum_{j=1}^{m}g_j}{m}
 =50-\frac{50}{m}n_{\mathrm{fail}},
\label{eq:fulfillment-base}
\end{equation}
where $n_{\mathrm{fail}}$ counts failed groups. A group earns its allocation only when all its items pass, irrespective of item counts or weights. Appendix~\ref{app:acceptance-groups} defines the CP/FN dimension scores.

Artifact Performance contributes a signed adjustment. Each criterion rating is mapped to points \emph{before} weighted averaging:
\begin{equation}
\begin{array}{c|rrrrrrr}
r & -3 & -2 & -1 & 0 & +1 & +2 & +3 \\
\phi(r) & -50 & -25 & -10 & 0 & +10 & +25 & +50
\end{array}
\label{eq:rating-points}
\end{equation}
For dimension $d\in\{C,O,V,D\}$ with criterion set $I_d$ and criterion weights $w_k$,
\begin{equation}
p_d=\frac{\sum_{k\in I_d}w_k\phi(r_k)}{\sum_{k\in I_d}w_k},
\qquad \Delta=\sum_{d\in\{C,O,V,D\}}\alpha_dp_d\in[-50,50],
\label{eq:quality}
\end{equation}
where $(\alpha_C,\alpha_O,\alpha_V,\alpha_D)=(0.25,0.15,0.50,0.10)$; Table~\ref{tab:weights} lists criterion weights. The total is
\begin{equation}
S=B+\Delta.
\label{eq:total-score}
\end{equation}
Thus $S\in[B-50,B+50]$, with an overall range of $[-50,100]$ and no clipping. Full fulfillment and ordinary performance yield 50; reaching 100 requires full fulfillment and maximal performance ratings.

We report the base $B$, CP and applicable FN passed-group rates, C/O/V/D point scores, adjustment $\Delta$, and total $S$ separately. Summaries use equal-task means and include task counts by family and purpose, along with completed, failed, and missing evaluations. Comparisons use common valid tasks under matching rubrics and scoring protocols; item-level evidence supports diagnosis beyond aggregate scores.

\section{Experiments}
\label{sec:experiments}

\subsection{Experimental Setup}
\label{sec:experimental-setup}
\paragraph{Tasks and agents.}
We evaluate nine model--harness configurations on the 211 VAS tasks: 47 presentations, 72 static graphics, 32 animations and videos, and 60 web interfaces. Each configuration receives the same task specifications and source materials, with evaluation rubrics withheld during authoring. All authoring configurations use their highest available reasoning setting. GPT-6 Astra, in a separate session with medium reasoning, serves as the common Agent Judge. Model identifiers, harness versions, reasoning labels, and detailed execution protocols are provided in the \href{https://github.com/Accio-Lab/VisualArtifactSpec}{released repository}.

\paragraph{Environment and protocol.}
Runs use isolated Accio Linux sandboxes with a 7,500-second task-stage limit. Submissions, rubrics, and judge instructions are pinned for each evaluation round. TF uses the CP/FN objective groups in revision \texttt{cp-fn-objectives-v2-20261114}, with all child checks required to pass and equal weights across groups. Website inspection covers both $1440\times900$ and $390\times844$ viewports and the required interactions; other artifacts follow the corresponding native-file, rendered, and temporal inspection requirements. Retained outputs are evaluated after authoring timeouts, and infrastructure or judge faults are handled separately from artifact defects.

\paragraph{Metrics.}
Following Section~\ref{sec:evaluation-protocol}, we report fulfillment base $B$, performance adjustment $\Delta$, and total score $S=B+\Delta$, together with Content, Organization, Visuals, and Delivery (C/O/V/D). Overall means weight tasks equally. Family profiles use the corresponding task subsets; purpose-level summaries cover the 151 non-web tasks with task-level purpose labels. Full-dataset analyses use all 1,899 configuration--task results; a fixed subset of 38 tasks illustrates task-level AP distributions for all nine configurations. Detailed numerical results are provided in Appendix~\ref{app:full-results}.

\begin{figure}[!t]
\centering
\includegraphics[width=\linewidth]{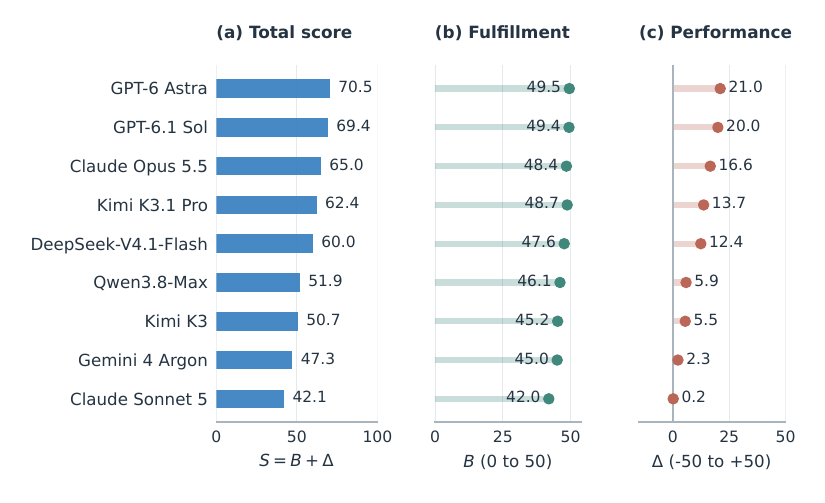}
\caption{Overall performance of nine authoring configurations. The panels show total score $S$, fulfillment base $B$, and performance adjustment $\Delta$, with explicit scales and a shared configuration order. Exact values are listed in Table~\ref{tab:main-results}.}
\label{fig:results-overall}
\end{figure}

\subsection{Overall Performance}
\label{sec:main-results}
\paragraph{Overall differences are concentrated in Artifact Performance.}
Figure~\ref{fig:results-overall} separates total scores into their two components. GPT-6 Astra and GPT-6.1 Sol score 70.52 and 69.37, respectively, followed by Claude Opus 5.5 at 65.03. Across configurations, mean $B$ ranges from 41.96 to 49.51, while mean $\Delta$ ranges from 0.16 to 21.01. Of the 28.40-point total gap between GPT-6 Astra and Claude Sonnet 5, 20.85 points (73.4\%) come from AP. Reporting the components separately therefore identifies whether a score difference primarily reflects unmet requirements or the quality of authoring choices.

\begin{figure}[!t]
\centering
\includegraphics[width=\linewidth]{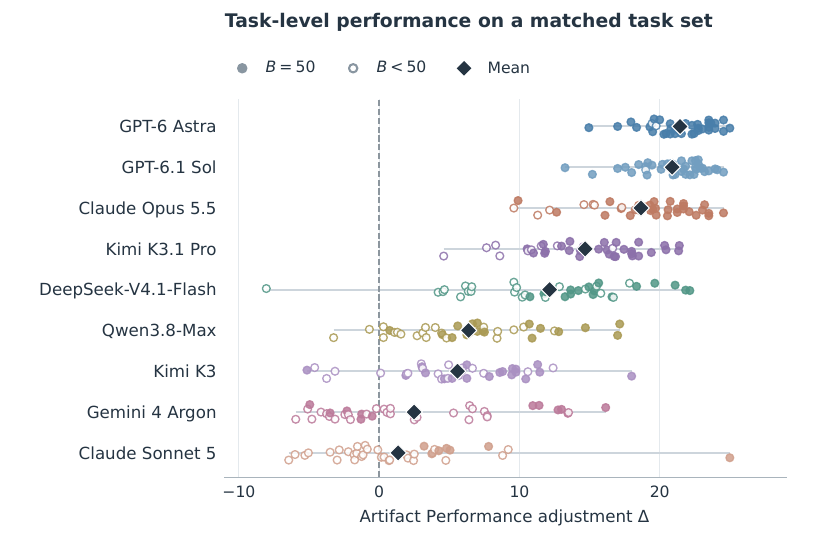}
\caption{Task-level AP on 38 matched tasks. Each circle represents one configuration--task result; filled circles have $B=50$, and open circles have $B<50$. Diamonds mark means, horizontal segments span the observed ranges, and the dashed line marks $\Delta=0$.}
\label{fig:results-task-distribution}
\end{figure}

\paragraph{Task Fulfillment and Artifact Performance reveal different limitations.}
Task-level results further show that similar TF scores can accompany different AP outcomes. Across the full dataset, outputs with $B=50$ span AP adjustments from $-5.55$ to $25.00$, revealing differences in authoring quality beyond requirement checks. Figure~\ref{fig:results-task-distribution} illustrates this distinction on a fixed display subset of 38 tasks.
Section~\ref{sec:case-studies} connects such differences to the underlying artifacts and task-specific criteria.

\subsection{Performance across Artifact Families and Contexts}
\label{sec:family-context-results}
\begin{figure}[!t]
\centering
\includegraphics[width=\linewidth]{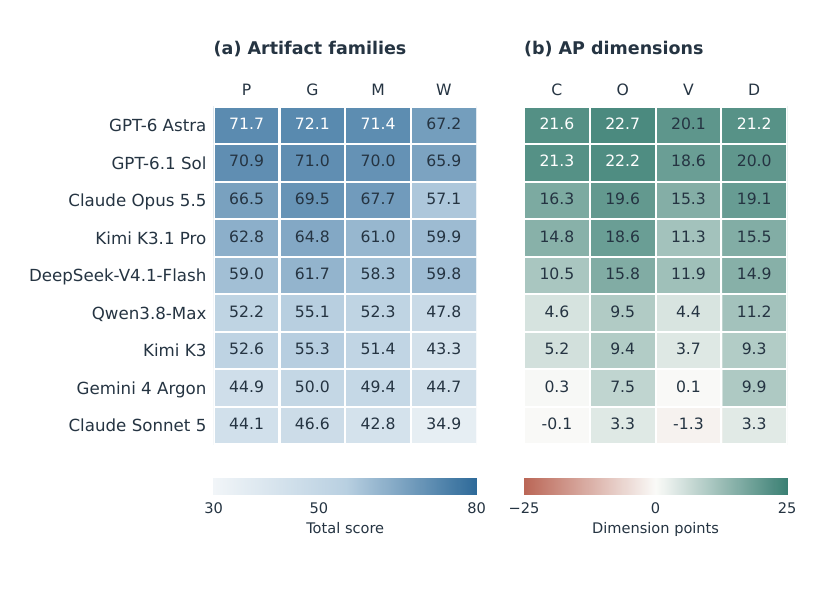}
\caption{Performance profiles by (a) artifact family and (b) AP dimension. P/G/M/W denote presentations, static graphics, animations/videos, and web interfaces; C/O/V/D denote Content, Organization, Visuals, and Delivery. Family cells show total scores. AP cells show dimension points before weighting, with a color scale centered at zero.}
\label{fig:results-profiles}
\end{figure}

\paragraph{Aggregate rankings conceal family-specific profiles.}
Figure~\ref{fig:results-profiles}(a) shows distinct profiles across artifact families. Claude Opus 5.5 scores 69.53 on static graphics but 57.05 on web interfaces; DeepSeek varies less, from 58.32 to 61.68 across the four families. GPT-6 Astra scores above 71 on presentations, static graphics, and animations/videos, compared with 67.20 on web interfaces. Cross-family evaluation therefore reveals both broad performance and strengths associated with particular deliverable forms.

\paragraph{Dimension profiles localize performance differences.}
Figure~\ref{fig:results-profiles}(b) shows that GPT-6 Astra exceeds 20 points in all four AP dimensions. Gemini 4 Argon scores 9.94 in Delivery but 0.33 in Content and 0.09 in Visuals; Claude Sonnet 5 has positive Organization and Delivery scores but negative Content and Visuals scores. An accessible deliverable and a followable structure can therefore coexist with weaknesses in content or visual execution.

Purpose-level results in Table~\ref{tab:purpose-results} further distinguish application settings. For example, DeepSeek scores 64.19 on promotion and engagement and 57.72 on practical guidance. These descriptive profiles combine differences in purpose, artifact family, and task difficulty.
The cases in Section~\ref{sec:case-studies} examine audience and medium adaptation through specific rubric judgments and visible evidence.

\subsection{Evaluation Analysis}
\label{sec:evaluation-analysis}
\label{sec:ablations}
\paragraph{Sensitivity to score aggregation.}
We vary three aggregation choices while holding artifacts and item judgments fixed: a linear AP map $50r/3$, equal C/O/V/D weights, and itemwise TF credit in place of grouped acceptance. Table~\ref{tab:aggregation-ablations} reports the results. Across the eight configurations compared under all three alternatives, the configuration ordering is preserved while score magnitudes change. Linear AP raises GPT-6 Astra by 7.31 points because it awards more points to $+1$ and $+2$ ratings. Equal AP weights raise Gemini 4 Argon by 2.22 points, reflecting its stronger Delivery and Organization scores. Itemwise TF raises Claude Sonnet 5 by 3.14 points by crediting passed checks within failed acceptance groups. These changes show how the aggregation rules control the contribution of intermediate quality ratings and incomplete acceptance objectives.

\paragraph{From aggregate scores to actionable evidence.}
Criterion-level statistics in Appendix~\ref{app:criterion-results} identify typography and legibility (V3) as the most frequent source of negative ratings, followed by traceability and handoff (D3), graphical craftsmanship (V6), and support for claims (C2).
Section~\ref{sec:case-studies} relates these differences to evidence grouping in a research presentation, mobile reading in educational cards, and marker--curve consistency in a teaching animation.

\begin{figure}[!t]
\centering
\includegraphics[width=\linewidth]{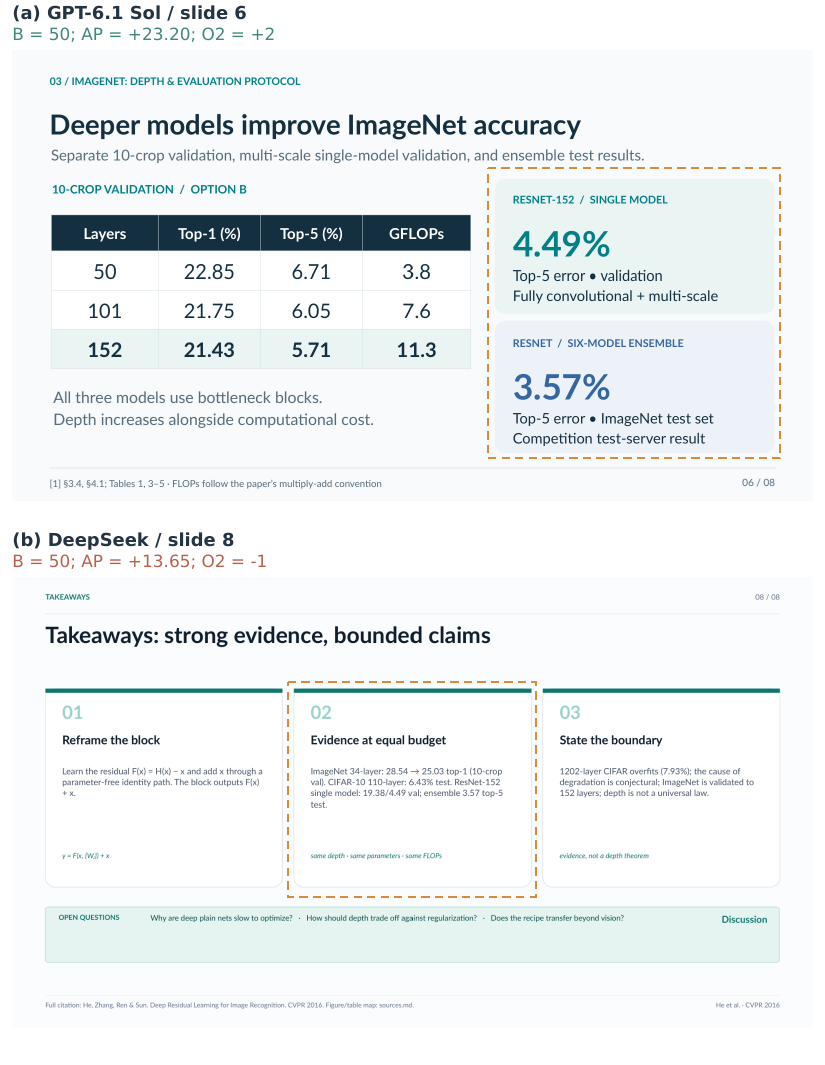}
\caption{Complete P01 slides with evidence groups marked by orange dashed boxes: separate protocol labels preserve the scope of reported results (a), whereas a shared qualifier spans heterogeneous comparisons (b).}
\label{fig:case-p01-evidence}
\end{figure}

\section{Case Studies}
\label{sec:case-studies}
\label{sec:capability-analysis}
We compare English-language artifacts from three tasks to relate recorded scores to visible authoring choices. Each pair receives the same fulfillment base; the cited AP criteria explain specific differences rather than the entire aggregate score gap.

\subsection{Keeping Evidence Conditions with Research Claims}
P01 requests an eight-slide ResNet presentation for a research-group meeting. GPT-6.1 Sol and DeepSeek both receive $B=50$, with AP adjustments of $23.20$ and $13.65$. Figure~\ref{fig:case-p01-evidence} shows how their slides organize experimental evidence. Sol separates single-model validation from ensemble test results and attaches the protocol and split to each number. DeepSeek groups results from different settings under ``Evidence at equal budget'' with a shared ``same depth / same parameters / same FLOPs'' qualifier. Recovering the qualifier's intended scope requires returning to an earlier slide. The local organization criterion O2 therefore assigns $+2$ and $-1$. For a research audience, preserving comparison conditions locally matters even when the requested results are present.

\begin{figure}[!t]
\centering
\includegraphics[width=\linewidth]{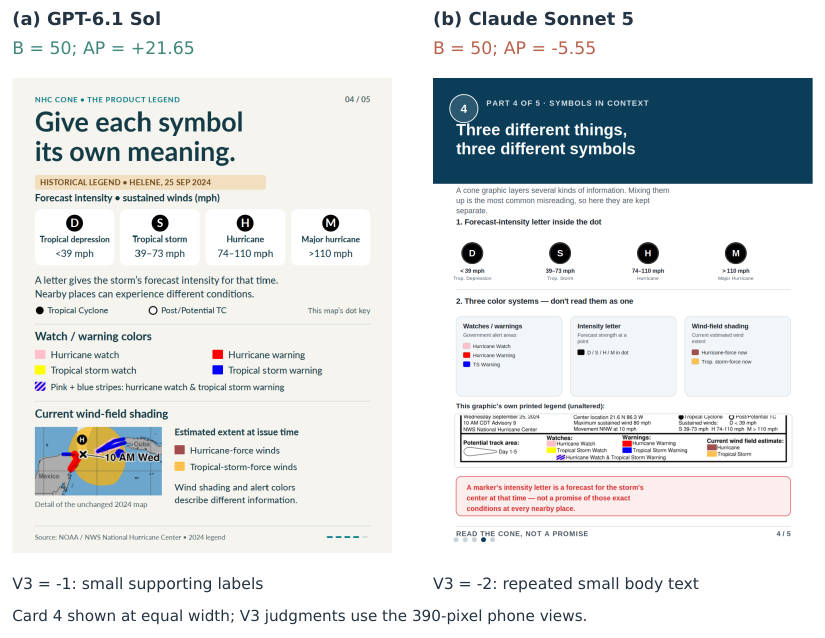}
\caption{P99: the same symbol lesson at equal display width. Phone-scale inspection distinguishes small supporting labels from repeated difficulty reading necessary explanations; AP summarizes all five cards.}
\label{fig:case-p99-mobile-comparison}
\end{figure}

\begin{figure}[!t]
\centering
\includegraphics[width=\linewidth]{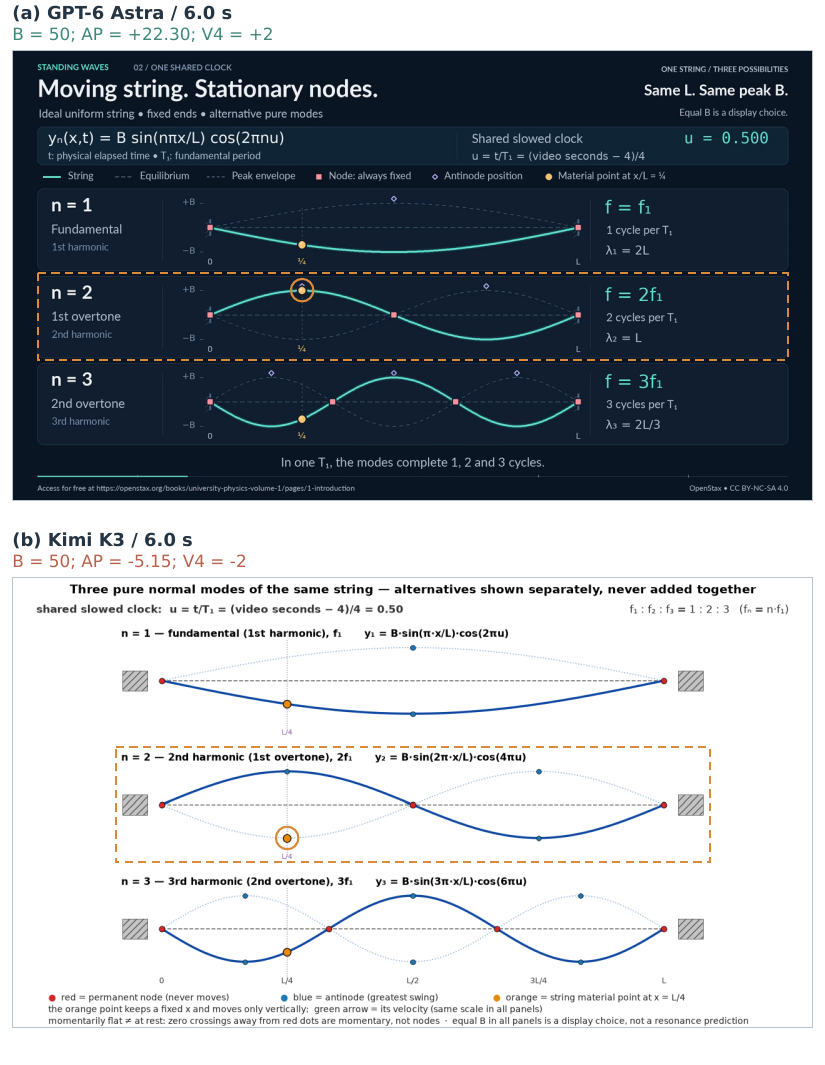}
\caption{Complete P121 frames at 6.0 seconds. Orange dashed boxes highlight the second mode, and circles mark the tracked material point: it follows the curve in Astra's frame but separates from it in Kimi's. Temporal evidence establishes the underlying rate mismatch.}
\label{fig:case-p121-motion-comparison}
\end{figure}

\subsection{Adapting Complete Information to Mobile Reading}
P99 requests five educational cards about hurricane forecast cones for a general audience reading on a phone. GPT-6.1 Sol and Claude Sonnet 5 both receive $B=50$, but their AP adjustments are $21.65$ and $-5.55$. Figure~\ref{fig:case-p99-mobile-comparison} compares their fourth cards at equal width. Sol separates the symbol classes into distinct bands with prominent definitions, while Sonnet distributes smaller explanations across a denser set of panels. At the prescribed 390-pixel inspection width, Sol's supporting labels remain small (V3: $-1$); Sonnet's necessary body and legend text repeatedly falls to roughly 5--8 pixels across the carousel (V3: $-2$). The comparison illustrates a medium-specific reading burden that is not resolved by including all required information, and also identifies a remaining weakness in the higher-scoring output.

\subsection{Maintaining Correspondence through Motion}
P121 requests a teaching animation of the first three standing-wave modes. Both artifacts receive $B=50$, with AP adjustments of $22.30$ for GPT-6 Astra and $-5.15$ for Kimi K3. In Figure~\ref{fig:case-p121-motion-comparison}, Astra's tracked material point remains on the second-mode curve, whereas Kimi's point lies below it at the same instant. The temporal inspection also finds that Kimi's three markers complete four cycles each over 4--20 seconds, although the curves complete four, eight, and twelve. This persistent mismatch yields $-2$ for both visual encoding (V4) and temporal rhythm (V8); Astra receives $+2$ and $+1$. The AP judgments reveal a temporal inconsistency not captured by the current TF checks, highlighting the need to verify marker--curve correspondence throughout the animation.

\section{Conclusion}
We introduced VisualArtifactSpec 1.0, a benchmark of 211 source-grounded creation and modification tasks spanning four artifact families and eight application purposes. VAS separates Task Fulfillment from Artifact Performance, using shared Content, Organization, Visuals, and Delivery criteria with task-specific rating anchors to assess authoring choices in context.

The results reveal substantial differences in Artifact Performance that Task Fulfillment alone does not capture. Family and purpose profiles reveal differences hidden by overall means, while artifact-based cases connect these differences to evidence organization, mobile readability, and temporal consistency. VAS thus provides a common framework for comparing authoring capabilities across artifact families and tracing performance limitations to concrete evidence. Future work should validate judge reliability and contextual ratings through human assessment and independent judges.

\clearpage
\bibliographystyle{plainnat}
\bibliography{references}
\clearpage
\appendix
\section{Task Classification and Deliverable Formats}
\label{app:scope}
\subsection{Classification Rules and Examples}
\label{app:classification}
Artifact families follow the principal deliverable objective, with secondary types listed in Table~\ref{tab:coverage}. Animated Graphics are standalone animated SVGs or looping GIFs; Animated Videos use authored diagrams, data graphics, or animation projects to produce a linear video; Edited Videos assemble supplied screenshots or media with timed focus, annotations, and captions. P25 is an Animated Video because it revises a re-renderable animation project, whereas P28 and P46 are Edited Videos that assemble interface screenshots into tutorials. Supplementary GIF previews do not create additional tasks.

For static graphics, multi-page series and explicit banner briefs take precedence; other single-page tasks distinguish focal-message posters from infographics and visual summaries, including evidence-led research posters.

\paragraph{Inputs and deliverable formats.}
Inputs include papers, reports, data, manuals, images, media, templates, and existing projects. Required deliverables include PPTX, Beamer source, HTML or Markdown presentation source, ODP, SVG and raster exports, animation files, and website projects. Each task specifies its required combination of native files and rendered outputs; input formats and supplementary previews do not determine the artifact family.

\paragraph{Task identifiers.}
The paper retains the original P-number shorthand for non-web tasks. Public identifiers preserve that number with an artifact-family prefix: for example, P80 is \texttt{presentations-080}, P99 is \texttt{static-graphics-099}, and P26 is \texttt{animations-videos-026}. Website identifiers run from \texttt{web-interfaces-001} to \texttt{web-interfaces-060} in sorted legacy-ID order. The accompanying \path{task-purpose-mapping.json} records all 211 identifier mappings and purpose assignments.

\paragraph{Application-purpose scope.}
Each task is assigned one primary purpose according to its principal application objective, with a specific use within that category. Table~\ref{tab:purpose-coverage} gives the distribution by artifact family. The 151 non-web assignments retain the established mapping; website assignments were reconstructed from task instructions and reconcile with the existing 60-task aggregate counts in Figure~\ref{fig:purposes}. The purpose-level results in Table~\ref{tab:purpose-results} retain the 151 non-web-task analysis subset.

\begin{table}[htbp]
\centering\small
\caption{Primary application purposes by artifact family. P: presentations; G: static graphics; M: animations/videos; W: web interfaces.}
\label{tab:purpose-coverage}
\begin{tabular}{@{}lrrrrr@{}}
\toprule Purpose & P & G & M & W & Total \\
\midrule
Research Communication & 18 & 6 & 0 & 0 & 24 \\
Decision Support & 14 & 2 & 0 & 19 & 35 \\
Conceptual Understanding & 9 & 21 & 26 & 5 & 61 \\
Practical Guidance & 3 & 14 & 3 & 4 & 24 \\
Promotion \& Engagement & 3 & 16 & 2 & 6 & 27 \\
Public Information & 0 & 13 & 1 & 4 & 18 \\
Culture \& Leisure & 0 & 0 & 0 & 8 & 8 \\
Production \& Operations & 0 & 0 & 0 & 14 & 14 \\
\midrule
Total & 47 & 72 & 32 & 60 & 211 \\
\bottomrule
\end{tabular}
\end{table}

\section{Rubric Details}
\label{app:weights}
\subsection{Dimension Scores and Exception Handling}
\label{app:acceptance-groups}
Acceptance groups follow objectives such as deliverable completeness, editability, source attribution, modification boundaries, accessibility, and required operations. Explicit page or card boundaries guide the refinement of coarse groups; otherwise, items are grouped by acceptance objective. Each group belongs to CP or FN, with membership frozen before agent comparisons. All child items must pass for a group to pass. Groups share the 50-point base equally, irrespective of their item counts; CP and FN have no separate fixed point budgets.

For dimension $d\in\{\mathrm{CP},\mathrm{FN}\}$ with nonempty group-index set $\mathcal{J}_d$, the dimension score is its passed-group percentage:
\begin{equation}
\mathrm{Score}_d=100\,\frac{\sum_{j\in\mathcal{J}_d}g_j}{|\mathcal{J}_d|}.
\end{equation}
FN is omitted when no functional groups apply. For example, with ten acceptance groups, each failed group deducts five points from the base regardless of the number of failed child items.

Aggregation requires valid memberships and complete, evidence-supported judgments. Missing required artifacts receive No on affected requirements when inspection establishes their absence; unresolved assessments remain unscored. A website-specific exception applies when evaluator faults have been excluded and no usable website exists because the submission is missing, contains only descriptions or screenshots, or cannot run due to its own defects: all CP/FN items receive No and all performance items receive $-3$. Local failures affect only dependent items. This exception does not apply to other artifact families or evaluator environment failures.

\subsection{Shared Performance Criteria}
\label{app:shared-criteria}
Table~\ref{tab:quality-criteria} defines the scopes of the 18 shared performance criteria. Each task instantiates these scopes through criterion wording, observation boundaries, and rating anchors tailored to its audience, purpose, and medium.
\begingroup
\small
\setlength{\tabcolsep}{4pt}
\renewcommand{\arraystretch}{1.12}
\setlength{\LTpre}{6pt}
\setlength{\LTpost}{10pt}
\begin{longtable}{@{}>{\raggedright\arraybackslash}p{0.04\linewidth}>{\raggedright\arraybackslash}p{0.30\linewidth}>{\raggedright\arraybackslash}p{\dimexpr0.66\linewidth-4\tabcolsep\relax}@{}}
\caption{Shared scopes of Artifact Performance criteria. Task-specific anchors determine how each scope is assessed for a particular artifact.}\label{tab:quality-criteria}\\
\toprule ID & Criterion & Evaluation scope \\
\midrule
\endfirsthead
\multicolumn{3}{c}{\small\itshape Table~\thetable{} (continued)}\\
\toprule ID & Criterion & Evaluation scope \\
\midrule
\endhead
\midrule
\multicolumn{3}{r}{\small\itshape Continued on next page}\\
\endfoot
\bottomrule
\endlastfoot
\multicolumn{3}{@{}l}{\textbf{Content}} \\*
C1 & Conceptual expression & Accurate, understandable concepts, terminology, qualifications, and references for the intended audience. \\
C2 & Support for claims & How evidence, data, or observed processes support the artifact's claims and reasoning. \\
C3 & Content selection & Relevance, priorities, information density, tone, and necessary qualifications for the intended purpose. \\
\midrule
\multicolumn{3}{@{}l}{\textbf{Organization}} \\*
O1 & Overall structure & The overall explanatory, argumentative, decision, or action path. \\
O2 & Local organization & Semantic order and relationships within a page, stage, node, or comparison unit. \\
O3 & Continuity and closure & Continuity across pages or states, or coherence and closure of branches within a single page. \\
\midrule
\multicolumn{3}{@{}l}{\textbf{Visuals}} \\*
V1 & Visual hierarchy & Entry points, focal content, and relative emphasis of primary and supporting information. \\
V2 & Composition and layout & Area allocation, grids, balance, spacing, and whitespace between information blocks. \\
V3 & Typography and legibility & Text visibility, size, contrast, spacing, and line breaks at the intended viewing scale. \\
V4 & Visual encoding & Recognizable marks and reliable correspondences between visual forms, meanings, and states. \\
V5 & Contextual aesthetics & Integration of visual language with the intended context, beyond color relationships alone. \\
V6 & Graphical craftsmanship & Contours, line widths, image crops, proportions, and local geometric or optical alignment. \\
V7 & Visual consistency & Stable identities, marks, styles, and content correspondences across pages, states, and formats. \\
V8 & Scanning and temporal rhythm & Eye movement, information reveal, dwell time, motion, and transitions. \\
V9 & Color relationships & Aesthetic coordination of hue, lightness, saturation, and color-area proportions. \\
\midrule
\multicolumn{3}{@{}l}{\textbf{Delivery}} \\*
D1 & Direct usability & Locating, opening, and directly using the required deliverables. \\
D2 & Editable structure & Editability and organization of required native objects, resources, and materials. \\
D3 & Traceability and handoff & Verifiability of required sources, calculations, changes, and usage instructions. \\
\end{longtable}
\endgroup

\subsection{Rating Evidence and Item Boundaries}
\label{app:rating-evidence}
Judgments identify the observed property and its location in rendered views, native-file extracts, temporal samples, or interaction records. Positive ratings explain the design choice and its benefit: $+1$ also identifies the remaining limitation; $+2$ requires inspection across the full applicable scope and active checks for weaknesses; $+3$ additionally demonstrates how a non-obvious choice resolves a task challenge beyond a conventional solution of comparable correctness and polish. Sampling coverage is retained with the evidence.

Each item is judged within its own scope. Shared evidence supports multiple ratings only when it establishes an impact within each item's boundaries. For example, reading contrast concerns V3, color-to-meaning correspondence concerns V4, and color harmony concerns V9. In modification tasks, evaluation covers permitted edits, their integration with retained content, and newly introduced defects; unchanged source defects and harmless style differences do not incur penalties.

\subsection{Criterion Weights}
Table~\ref{tab:weights} gives the weights used in Equation~\ref{eq:quality}. Each rating is mapped through $\phi$ before weighted averaging within its dimension. Thus, equally weighted ratings of 0 and $+2$ yield 12.5 points, rather than the 10 points obtained by mapping their mean rating. The criterion weights sum to 100, with dimension totals matching the C/O/V/D aggregation weights. The versioned scoring implementation and score-reporting conventions are available in the \href{https://github.com/Accio-Lab/VisualArtifactSpec}{released repository}.
\begin{table}[H]
\centering\small
\caption{Criterion and dimension weights.}
\label{tab:weights}
\begin{tabular}{@{}llr@{}}
\toprule Dimension & Criterion weights in order & Total weight \\
\midrule Content (C1--C3) & 9, 9, 7 & 25\% \\
Organization (O1--O3) & 6, 5, 4 & 15\% \\
Visuals (V1--V9) & 5, 8, 7, 5, 7, 6, 3, 3, 6 & 50\% \\
Delivery (D1--D3) & 4, 3, 3 & 10\% \\
\bottomrule
\end{tabular}
\end{table}

\FloatBarrier
\section{Benchmark Comparison Definitions and Evidence}
\label{app:benchmark-comparison}
Table~\ref{tab:benchmark-comparison-candidate} compares authoring evaluation protocols, including generation, editing, and reconstruction. For evaluation criteria, a check denotes routine assessment through a shared rubric or recurring checks across relevant task categories; isolated examples are insufficient. A dash denotes that routine assessment is not established. The TF/AP grouping organizes this comparison and need not reproduce a benchmark's original scoring structure.

\subsection{Coverage and Evaluation Criteria}
Coverage follows the evaluated deliverable: P denotes complete presentation decks, G static graphics, M animations/videos, and W interactive web interfaces. Slide images and single-slide reconstruction do not establish P; static webpage renderings do not establish W.

\paragraph{Task Fulfillment.}
Instruction compliance checks explicit content, structure, modification, and deliverable obligations. Functional verification checks operations and their resulting outputs or state transitions; rendering success alone is insufficient.

\paragraph{Artifact Performance.}
Visual presentation covers appearance, layout, and legibility. Content accuracy covers factual and quantitative correctness, consistency of semantic relationships, and fidelity to source or reference content; visual resemblance alone is insufficient. Audience alignment assesses fit to readers' prior knowledge and needs. Medium adaptation assesses fit to a medium or platform's usage preferences through choices about information density, organization, pacing, and visual presentation. Prompt-level audience descriptions, responsive resizing, or general legibility alone do not establish these last two criteria.

These AP columns support cross-benchmark comparison. Content accuracy is assessed within VAS's Content dimension; audience alignment and medium adaptation are expressed through task-specific C/O/V criteria. Organization and Delivery remain distinct VAS dimensions but are not independent columns in Table~\ref{tab:benchmark-comparison-candidate}.

\subsection{Evidence for Table Entries}
Table~\ref{tab:comparison-evidence} records the principal evaluation evidence and relevant scope boundaries. Paper section and table numbers refer to the cited benchmark reports. Released checklists provide additional evidence for \href{https://huggingface.co/datasets/PresentBench/PresentBench/tree/main}{PresentBench} and \href{https://huggingface.co/datasets/tencent/ArtifactsBenchmark/blob/main/artifacts_bench.json}{ArtifactsBench}.

\begingroup
\small
\setlength{\tabcolsep}{4pt}
\renewcommand{\arraystretch}{1.12}
\setlength{\LTpre}{6pt}
\setlength{\LTpost}{10pt}
\begin{longtable}{@{}>{\raggedright\arraybackslash}p{0.25\linewidth}>{\raggedright\arraybackslash}p{\dimexpr0.75\linewidth-2\tabcolsep\relax}@{}}
\caption{Evaluation evidence supporting the benchmark comparison.}\label{tab:comparison-evidence}\\
\toprule Benchmark & Evidence and scope \\
\midrule
\endfirsthead
\multicolumn{2}{c}{\small\itshape Table~\thetable{} (continued)}\\
\toprule Benchmark & Evidence and scope \\
\midrule
\endhead
\midrule
\multicolumn{2}{r}{\small\itshape Continued on next page}\\
\endfoot
\bottomrule
\endlastfoot
StructBench \citep{zhuo2026structbench} & Atomic factual questions support content accuracy; recurring appearance and spatial-arrangement checks support visual presentation. \\
\addlinespace
DrawAI-Bench \citep{cao2026drawai} & Fidelity criteria compare text content, formula relations, and connector semantics with references (Appendix~A, Table~12), supporting content accuracy alongside visual and editability checks. \\
\addlinespace
BizGenEval \citep{li2026bizgeneval} & Text and knowledge checks assess textual and factual correctness across visual domains, alongside visual and semantic constraint checks for instruction compliance (Sec.~3.1--3.4). \\
\addlinespace
VAO-Bench \citep{guo2026vao} & Content preservation, dependent-update checks, and domain-validity ratings assess content consistency alongside visual and task-requirement checks (Appendix~B.3). Its slide and webpage images count as G. \\
\addlinespace
PresentBench \citep{chen2026presentbench} & Material-dependent checklists assess factual and quantitative fidelity; shared rubrics assess visual design. Audience checks cover explanation difficulty and instructional goals across presentation contexts; medium-specific reading practices are not a routine separate target. \\
\addlinespace
Code2Video \citep{chen2025code2video} & The Accuracy \& Depth criterion assesses the correctness of presented knowledge alongside visual quality and temporal logic (Sec.~3.3). TeachQuiz measures knowledge transfer; this alone does not establish audience or medium adaptation. \\
\addlinespace
Vision2Web \citep{he2026vision2web} & Required interactions and resulting states support functional verification. Multi-resolution prototype comparisons assess responsive visual fidelity (Sec.~2.1, 3.3). \\
\addlinespace
EvoPresent \citep{liu2025evopresent} & Deck and video evaluation covers source fidelity, visual design, and narrative coherence. Its human checklist includes distance legibility (Appendix~E); source fidelity alone does not establish instruction compliance. \\
\addlinespace
GDB \citep{deganutti2026graphicdesignbench} & Text, typography, preservation, and requested-motion checks support compliance (Sec.~7.5, Table~44). Font-size hierarchy and general M-Judge quality criteria (Table~65) do not establish routine content-accuracy or audience-fit assessment. \\
\addlinespace
ArtifactsBench \citep{zhang2025artifactsbench} & Scripted interactions and task checklists assess compliance, function, and visuals. Recurring checks of factual, quantitative, and conceptual content support content accuracy; device compatibility and occasional audience-specific checks do not establish the last two columns. \\
\end{longtable}
\endgroup

\paragraph{VAS task-specific anchors.}
P80's research-group slides are assessed for a focal claim on each projected page and pauses between pages (V1, V8). P99's social-media carousel is assessed for card-level entry points, information density, scanning sequence, and pacing across five cards (V8). These instantiate shared criteria for different readers and presentation practices.
\FloatBarrier

\section{Supplementary Experimental Results}
\label{app:full-results}
\subsection{Overall and Family Scores}
Tables~\ref{tab:main-results} and~\ref{tab:family-results} give the exact values underlying Figures~\ref{fig:results-overall} and~\ref{fig:results-profiles}. Task-level scores and scripts for reproducing the statistical figures and aggregate tables accompany the source release.

\begin{table}[!htbp]
\centering\small
\setlength{\tabcolsep}{4pt}
\renewcommand{\arraystretch}{1.1}
\caption{Overall scores and Artifact Performance dimensions. $B$ is the fulfillment base, $\Delta$ the performance adjustment, and $S=B+\Delta$ the total. C/O/V/D are dimension point scores in $[-50,50]$, before dimension weighting; higher is better.}
\label{tab:main-results}
\begin{tabularx}{\linewidth}{@{}Yrrrrrrr@{}}
\toprule
\textbf{Configuration} & $B$ & $\Delta$ & $S$ & C & O & V & D \\
\midrule
GPT-6 Astra & 49.51 & 21.01 & 70.52 & 21.61 & 22.75 & 20.14 & 21.22 \\
GPT-6.1 Sol & 49.40 & 19.98 & 69.37 & 21.30 & 22.23 & 18.63 & 20.00 \\
Claude Opus 5.5 & 48.44 & 16.59 & 65.03 & 16.33 & 19.61 & 15.31 & 19.14 \\
Kimi K3.1 Pro & 48.72 & 13.67 & 62.38 & 14.77 & 18.57 & 11.27 & 15.54 \\
DeepSeek-V4.1-Flash & 47.61 & 12.42 & 60.03 & 10.51 & 15.77 & 11.87 & 14.86 \\
Qwen3.8-Max & 46.07 & 5.87 & 51.95 & 4.58 & 9.53 & 4.36 & 11.19 \\
Kimi K3 & 45.21 & 5.49 & 50.70 & 5.20 & 9.37 & 3.70 & 9.34 \\
Gemini 4 Argon & 45.02 & 2.25 & 47.28 & 0.33 & 7.54 & 0.09 & 9.94 \\
Claude Sonnet 5 & 41.96 & 0.16 & 42.12 & -0.08 & 3.32 & -1.29 & 3.27 \\
\bottomrule
\end{tabularx}
\end{table}

\begin{table}[!htbp]
\centering\small
\setlength{\tabcolsep}{4pt}
\renewcommand{\arraystretch}{1.1}
\caption{Total scores by artifact family. Parentheses give task counts; overall scores weight tasks equally across the four families.}
\label{tab:family-results}
\begin{tabularx}{\linewidth}{@{}Yrrrr@{}}
\toprule
\textbf{Configuration} & \shortstack{Presentations\\(47)} & \shortstack{Static graphics\\(72)} & \shortstack{Animations/videos\\(32)} & \shortstack{Web interfaces\\(60)} \\
\midrule
GPT-6 Astra & 71.73 & 72.10 & 71.41 & 67.20 \\
GPT-6.1 Sol & 70.92 & 70.96 & 69.98 & 65.94 \\
Claude Opus 5.5 & 66.51 & 69.53 & 67.71 & 57.05 \\
Kimi K3.1 Pro & 62.83 & 64.75 & 60.96 & 59.95 \\
DeepSeek-V4.1-Flash & 58.97 & 61.68 & 58.32 & 59.79 \\
Qwen3.8-Max & 52.15 & 55.12 & 52.26 & 47.81 \\
Kimi K3 & 52.61 & 55.35 & 51.38 & 43.28 \\
Gemini 4 Argon & 44.90 & 50.05 & 49.36 & 44.71 \\
Claude Sonnet 5 & 44.13 & 46.55 & 42.83 & 34.85 \\
\bottomrule
\end{tabularx}
\end{table}

\subsection{Application-Purpose Scores}
Table~\ref{tab:purpose-results} reports the 151 individually mapped non-web tasks across six purposes (Appendix~\ref{app:scope}). Each column describes its task subset; artifact-family composition and difficulty also vary across columns.

\begin{table}[!htbp]
\centering\small
\setlength{\tabcolsep}{4pt}
\renewcommand{\arraystretch}{1.1}
\caption{Total scores by application purpose for the 151 non-web tasks. Columns denote research communication, decision support, conceptual understanding, practical guidance, promotion and engagement, and public information; parentheses give task counts.}
\label{tab:purpose-results}
\begin{tabularx}{\linewidth}{@{}Yrrrrrr@{}}
\toprule
\textbf{Configuration} & \shortstack{Research\\(24)} & \shortstack{Decision\\(16)} & \shortstack{Concepts\\(56)} & \shortstack{Guidance\\(20)} & \shortstack{Promotion\\(21)} & \shortstack{Public info.\\(14)} \\
\midrule
GPT-6 Astra & 71.89 & 72.52 & 71.96 & 71.76 & 70.49 & 72.62 \\
GPT-6.1 Sol & 71.40 & 71.04 & 70.88 & 69.55 & 69.80 & 71.78 \\
Claude Opus 5.5 & 65.20 & 67.92 & 68.53 & 68.72 & 69.21 & 70.09 \\
Kimi K3.1 Pro & 61.64 & 65.15 & 63.08 & 61.92 & 64.11 & 66.24 \\
DeepSeek-V4.1-Flash & 57.45 & 59.82 & 59.90 & 57.72 & 64.19 & 63.30 \\
Qwen3.8-Max & 51.43 & 53.09 & 53.85 & 50.59 & 56.03 & 57.46 \\
Kimi K3 & 51.61 & 55.58 & 53.41 & 50.45 & 56.61 & 56.07 \\
Gemini 4 Argon & 43.73 & 44.96 & 49.47 & 45.71 & 54.59 & 49.54 \\
Claude Sonnet 5 & 42.01 & 47.39 & 45.28 & 40.60 & 48.25 & 47.78 \\
\bottomrule
\end{tabularx}
\end{table}

\subsection{Scoring Sensitivity}
\label{app:scoring-sensitivity}
Table~\ref{tab:aggregation-ablations} isolates three aggregation choices while holding artifacts, item judgments, and all remaining scoring components fixed.

\begin{table}[!htbp]
\centering\small
\setlength{\tabcolsep}{4pt}
\renewcommand{\arraystretch}{1.1}
\caption{Scoring sensitivity with fixed artifacts and item judgments. Linear AP replaces the nonlinear rating map; equal AP uses four equal dimension weights; itemwise TF replaces acceptance-group scoring with the fraction of passed binary items. Dashes denote unreported comparisons.}
\label{tab:aggregation-ablations}
\begin{tabularx}{\linewidth}{@{}Yrrrr@{}}
\toprule
\textbf{Configuration} & Full protocol & Linear AP & Equal AP & Itemwise TF \\
\midrule
GPT-6 Astra & 70.52 & 77.83 & 70.94 & 70.72 \\
GPT-6.1 Sol & 69.37 & 76.40 & 69.94 & 69.63 \\
Claude Opus 5.5 & 65.03 & 71.07 & 66.04 & 65.61 \\
Kimi K3.1 Pro & 62.38 & --- & 63.75 & --- \\
DeepSeek-V4.1-Flash & 60.03 & 64.51 & 60.87 & 61.21 \\
Qwen3.8-Max & 51.95 & 53.85 & 53.49 & 53.68 \\
Kimi K3 & 50.70 & 52.40 & 52.12 & 52.74 \\
Gemini 4 Argon & 47.28 & 47.80 & 49.50 & 49.57 \\
Claude Sonnet 5 & 42.12 & 41.76 & 43.27 & 45.26 \\
\bottomrule
\end{tabularx}
\end{table}

\subsection{Matched Task Set}
\label{app:matched-tasks}
Figure~\ref{fig:results-task-distribution} displays a fixed subset of 38 tasks shared by all nine configurations; the main findings use all 211 tasks. Table~\ref{tab:matched-task-ids} lists their IDs. The panel contains 342 configuration--task results. Jitter separates points, and horizontal lines show observed ranges.

\begin{table}[!htbp]
\centering\small
\caption{The fixed task set used for the paired AP distributions.}
\label{tab:matched-task-ids}
\begin{tabularx}{\linewidth}{@{}Y@{}}
\toprule Task IDs \\
\midrule
P01, P02, P05, P07, P09, P10, P12, P17, P18, P20, P22, P23, P24, P26, P28, P29, P31, P33, P35, P36, P37, P112, P121, P124, P139, P140, P148, P152, P164, P183, P197, P199, P202, P207, P212, P215, P216, P219 \\
\bottomrule
\end{tabularx}
\end{table}

\subsection{Criterion-Level Error Patterns}
\label{app:criterion-results}
Criterion-level diagnostics use 1,688 assessments from eight configurations: GPT-6 Astra, GPT-6.1 Sol, Claude Opus, DeepSeek, Qwen, Kimi K3, Gemini, and Claude Sonnet. Negative-rating rates are averaged over tasks within each configuration and then equally across configurations. The most frequent negative ratings occur for typography and legibility (V3, 52.31\%), traceability and handoff (D3, 26.72\%), graphical craftsmanship (V6, 25.06\%), and support for claims (C2, 24.88\%). Direct usability (D1) receives negative ratings in 2.43\% of assessments. These rates distinguish the ability to deliver an accessible artifact from the readability, interpretability, and evidential support of its details.

\subsection{Additional Website Evidence}
Kimi K3's CON-002 reading website receives $B=30.77$ and $\Delta=4.65$. Two recorded interaction sequences illustrate failed functional checks:

\begin{table}[H]
\centering\small
\caption{Required state transitions and observed failures in CON-002.}
\label{tab:website-failure-evidence}
\begin{tabularx}{\linewidth}{@{}p{0.12\linewidth}YY@{}}
\toprule Criterion & Action and expected result & Observed result \\
\midrule
FN01 & With a saved reading position, select a search result labeled as the article title; navigate to that title. & The interface restores the old reading position at scroll offset 1350, leaving the title above the viewport. \\
\addlinespace
FN10 & After a storage-write failure, restore write access, add a bookmark, and reload; retain the new bookmark. & The interface confirms addition, but the persistent store retains only the earlier bookmarks; new entries disappear after reload. \\
\bottomrule
\end{tabularx}
\end{table}
These observations test navigation and recovery across states, beyond the presence of search and bookmark controls.

\end{document}